\documentclass[10pt,twocolumn,letterpaper]{article}

\usepackage[pagenumbers]{cvpr} 

\usepackage{graphicx}
\usepackage{amsmath}
\usepackage{amssymb}
\usepackage{booktabs}
\usepackage{amsmath}
\usepackage{multirow}
\usepackage{array}
\usepackage{makecell}

\usepackage{pifont}
\newcommand{\cmark}{\ding{51}}%
\usepackage[pagebackref,breaklinks,colorlinks]{hyperref}

\usepackage[capitalize]{cleveref}
\crefname{section}{Sec.}{Secs.}
\Crefname{section}{Section}{Sections}
\Crefname{table}{Table}{Tables}
\crefname{table}{Tab.}{Tabs.}

\def\cvprPaperID{1109} 
\def\confName{CVPR}
\def\confYear{2022}

\begin{document}

\title{MonoJSG: Joint Semantic and Geometric Cost Volume for \\Monocular 3D Object Detection}

\author{Qing Lian \textsuperscript{1},~~~ Peiliang Li\textsuperscript{2},~~~ Xiaozhi Chen\textsuperscript{2} \\
\textsuperscript{1}The Hong Kong University of Science and Technology, 
\textsuperscript{2}DJI\\
{\tt\small qlianab@connect.ust.hk, peiliang.uav@gmail.com, cxz.thu@gmail.com}
}
\maketitle

\begin{abstract}
     Due to the inherent ill-posed nature of 2D-3D projection, monocular 3D object detection lacks accurate depth recovery ability. Although the deep neural network (DNN) enables monocular depth-sensing from high-level learned features, the pixel-level cues are usually omitted due to the deep convolution mechanism. To benefit from both the powerful feature representation in DNN and pixel-level geometric constraints, we reformulate the monocular object depth estimation as a progressive refinement problem and propose a joint semantic and geometric cost volume to model the depth error. 
     Specifically, we first leverage neural networks to learn the object position, dimension, and dense normalized 3D object coordinates. Based on the object depth, the dense coordinates patch together with the corresponding object features is reprojected to the image space to build a cost volume in a joint semantic and geometric error manner. The final depth is obtained by feeding the cost volume to a refinement network, where the distribution of semantic and geometric error is regularized by direct depth supervision.
    Through effectively mitigating depth error by the refinement framework, we achieve state-of-the-art results on both the KITTI and Waymo datasets.\footnote{Code available at \scriptsize{\url{https://github.com/lianqing11/MonoJSG}}}
    

\end{abstract}
\begin{figure}
    \centering
    \includegraphics[width = 0.49\textwidth, trim = 0 0 0 0, clip]{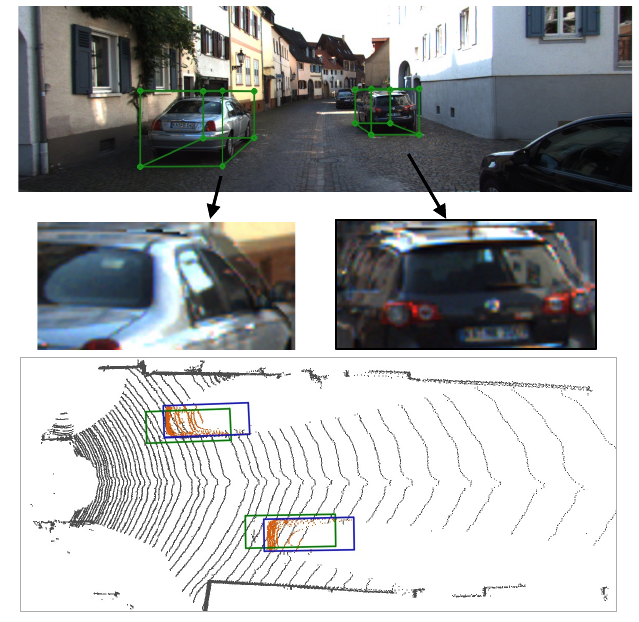}
    \caption{From top to bottom: Visualization of the estimated corners in the image space, which can be used to constrain an initial 3D bounding box;
     Reproject the object patch to the original image using the initial bounding boxes; Initial (\textcolor{green}{green}) and our MonoJSG refined (\textcolor{blue}{blue}) bounding boxes in the BEV space respectively. Compared with corners, photometric mismatch provides discriminative features for identifying localization error.
     }
    \label{fig:fig1}
\end{figure}

\section{Introduction}
\label{sec:intro}

As a fundamental component in 3D perception, 3D object detection has drawn increasing attention from the area of autonomous driving, robotic navigation, \textit{etc}.
Recently, it has achieved remarkable progress based on lidar or stereo sensing solutions. 
However, the high cost of lidar sensors and the complicated  online calibration in stereo cameras limit their mass applications in downstream tasks.
Therefore, researchers start to focus on a cheaper alternative, monocular-based sensing solution.
Yet due to the ill-posed 2D-3D projection, the localization accuracy of monocular 3D object detection is far behind the lidar and stereo-based approaches.

Driven by powerful neural networks, multiple approaches~\cite{zhou2019objects, brazil2019m3d, pseudo_lidar, RTM3D, monodle, MonoFlex} are proposed to alleviate the challenging monocular depth recovery from different perspectives.
From the perspective of data formation, pseudo-lidar based approaches~\cite{pseudo_lidar, you2019pseudo, Ma_2019_ICCV, ma2020rethinking} transform the input images to pseudo point cloud and directly adopt lidar detectors on it. Although they achieve better performance over traditional approaches~\cite{chen2016monocular, mousavian20173d}, the heavy reliance on the depth prediction network leads to high latency and overfitting~\cite{Simonelli_2021_ICCV_pseudo}.
From the perspective of geometry reasoning, geometric constraint based approaches~\cite{RTM3D, MonoFlex} leverage neural networks to predict variant 2D cues then solve the object depth according to 2D-3D projection constraints. 
In particular, the 2D-3D constraints are built from the object edges~\cite{MonoFlex, decomp_shi, mousavian20173d, GuPNet}, sparse keypoints~\cite{RTM3D, autoshape_liu, km3dnet}, dense keypoints~\cite{MonoRun}, \textit{etc}.
Although reasoning depth by 2D-3D constraints is interpretable and easy-to-trace, the optimization gap between the indirect 2D cue prediction and direct depth prediction limits the final depth solving accuracy, \textit{i.e.,} the overall minimum 2D loss on whole training data does not necessarily mean the best depth estimation performance. 
As visualized in Figure~\ref{fig:fig1}, although the estimated bounding boxes' edges and corners look almost accurate in the image, the solved 3D bounding box has a non-trivial localization error from the bird's eye view. This localization error is agnostics from regressed 2D cues. However, if we reproject the object to the original image using the inaccurate location, a significant photometric misalignment can be observed in Figure~\ref{fig:fig1}, which inspires our joint semantic and geometric depth refinement approach.

In this work, we propose an approach called Joint Semantic and Geometric Cost Volume (MonoJSG), which utilizes pixel-level visual cues to refine bounding box proposals.
We first enrich the traditional 2D-3D constraint~\cite{RTM3D, autoshape_liu} by extra estimating the location of each pixel in the normalized object coordinate. Based on the estimated object depth, the normalized object coordinate is projected into the image space to build a pixel-level constraint for each bounding box. 
The pixel-level constraint measures the geometric error between each pixel's 2D location and the projection location of the normalized object coordinate. We further enrich the constraints with a semantic error, which measures the distance of the features queried by the origin 2D location and the projection location. 
As Figure~\ref{fig:fig1} shows, although the depth error can be obviously revealed by pixel-level raw photometric error, we found that simply extending this strategy to all instances cannot always refine accurate depth due to the variant textureless and irregular regions (\textit{e.g.,} the windshield or the rear window, \textit{etc.}). 
We instead leverage neural networks to learn semantic features as a more robust and discriminative representation compared with the raw image intensity.
Based on the designed joint geometric and semantic error manner, we construct a 4D cost volume to draw the error distribution around proposal depth for refinement. To make the cost volume adapt to variant depth error, its size is customized for each proposal based on a predicted depth uncertainty. 
Then a refinement network is designed to take the adaptive cost volume as input and output the final depth. 

Our approach shows benefits from two perspectives. From the view of explicit constraining, the exploited semantic features provide more dense cues to measure the localization error than pure sparse keypoints.
From the view of data-driven, we force the network to learn discriminative features that are suitable for refinement by end-to-end depth supervision.
With the aforementioned advantages, the proposed framework achieves superior performance, leading to new state-of-the-arts on the KITTI and Waymo datasets.

We summarize our main contributions as follows:
\begin{itemize}
\item Based on pixel-level geometric and semantic visual cues, we present a novel joint semantic and geometric error measuring approach for object depth. 

\item We design an adaptive 4D cost volume that models the error distribution for depth refinement. 
\item We demonstrate the effectiveness of the proposed approach on both the KITTI and Waymo datasets, which achieve state-of-the-art results with real-time performance. 
\end{itemize}

\section{Related work}
\subsection{Image-based Monocular detection}
The objective of monocular 3D object detection is to identify objects of interest and localize their 3D bounding boxes from a single image. To alleviate the ambiguity 2D-3D projection problem, existing approaches either leverage neural networks to extract high-level semantic representation, design 2D-3D geometry constraints, or incorporating external depth information for depth reasoning. 

\noindent{\textbf{Semantic representation based}}
Some representative approaches~\cite{chen20153d, chen2016monocular, simonelli2019disentangling,zhou2019objects, brazil2019m3d, manhardt2019roi} train neural networks to learn semantic representation and then directly regress the 3D bounding boxes based on the learned representation. Later work exploits from the perspective of network architecture~\cite{brazil2019m3d, CaDDN, M3DSSD}, objective function~\cite{simonelli2019disentangling, GuPNet}, \textit{etc}. CenterNet~\cite{zhou2019objects} proposes a centerness-based object detection paradigm and lifts the 2D detector to 3D space by adding several 3D task heads. M3D-RPN proposes a 3D-anchor that aligns the 2D anchors with 3D statistics. Shi et al.~\cite{decomp_shi} and Lu et al.~\cite{GuPNet} utilize geometric priors to decompose the location estimation into 2D and 3D height estimation. MonoFlex~\cite{MonoFlex} proposes a disentangled network to handle the bounding boxes with different degrees of truncation. However, as discussed in~\cite{monodle, MonoFlex}, the learned semantic representation is unexplainable and easy to overfit on some spurious features in the training data. 

\noindent{\textbf{Geometry constraint based}}
This line of approaches~\cite{mousavian20173d, RTM3D, li2019stereo, km3dnet, MonoRun, autoshape_liu, monopair} reason 3D location with 2D-3D geometric constraints. Mousavian et al.~\cite{mousavian20173d} first attempt to recover the 3D location by solving the constraints between 2D bounding boxes edges and 3D dimension. Li et al.~\cite{RTM3D, km3dnet} and Liu et al.~\cite{autoshape_liu} propose a keypoint based approaches to further limit the searching space of the geometric constraints in~\cite{mousavian20173d}.
MonoRun~\cite{MonoRun} proposes a self-supervised algorithm to learn a pixel-level constraint and recovers the 3D location by adopting a modified PnP solver.
However, the connection between the keypoint localization accuracy and depth error is indirect, where the optimal keypoint localization model does not guarantee the minimal depth error. Such indirect connection introduces multiple failure cases in this line of approach, \textit{i.e.,} the far away objects have small localization error but large depth estimation deviation. Furthermore, the semantic information in the geometric constraint based approach is under-exploited, which can provide a strong visual cue for depth reasoning. 

\noindent{\textbf{Depth-assisted monocular detection}}
Instead of directly taking images as the input of neural networks, depth-assisted based approaches convert the RGB image to a dense depth map and then utilize it to assist the recovery of 3D information. Pseudo lidar based approaches~\cite{pseudo_lidar, you2019pseudo, Ma_2019_ICCV, ma2020rethinking, pct} convert the depth map into pseudo point cloud and adopt point cloud based detectors to localize 3D bounding boxes. Instead of transforming the depth map into point cloud, other approaches~\cite{d4lcn, DDMP, GuPNet} leverage the depth map to guide the learning of 2D convolution. 
Although the depth-assisted based approaches achieve better performance, they are required to train an extra depth estimation network, which often needs more training data and is inefficient in inference. 

\subsection{Cost volume for 3D representation}
In the area of stereo matching~\cite{psmnet, groupwisestereo, pwcnet}, multi-view stereo~\cite{mvsnet, point-mvsnet}, \textit{etc}, cost volume is a widely adopted technique to compute the matching cost. Stereo matching approaches~\cite{psmnet, groupwisestereo, pwcnet} leverage a siamese network to extract features from the left and right cameras and then apply correlation-based or concatenation-based cost volume to compute the matching cost. In multi-view stereo, MVSNet~\cite{mvsnet} generates depth map by constructing a plane-sweep volume in the camera frustum space. However, the application of cost volume in monocular 3D object detection is not fully exploited. To the best of our knowledge, we propose the first 2D-3D cost volume that computes the matching cost for object depth.  
\begin{figure*}[!htb]
    \centering
    \includegraphics[width = \textwidth]{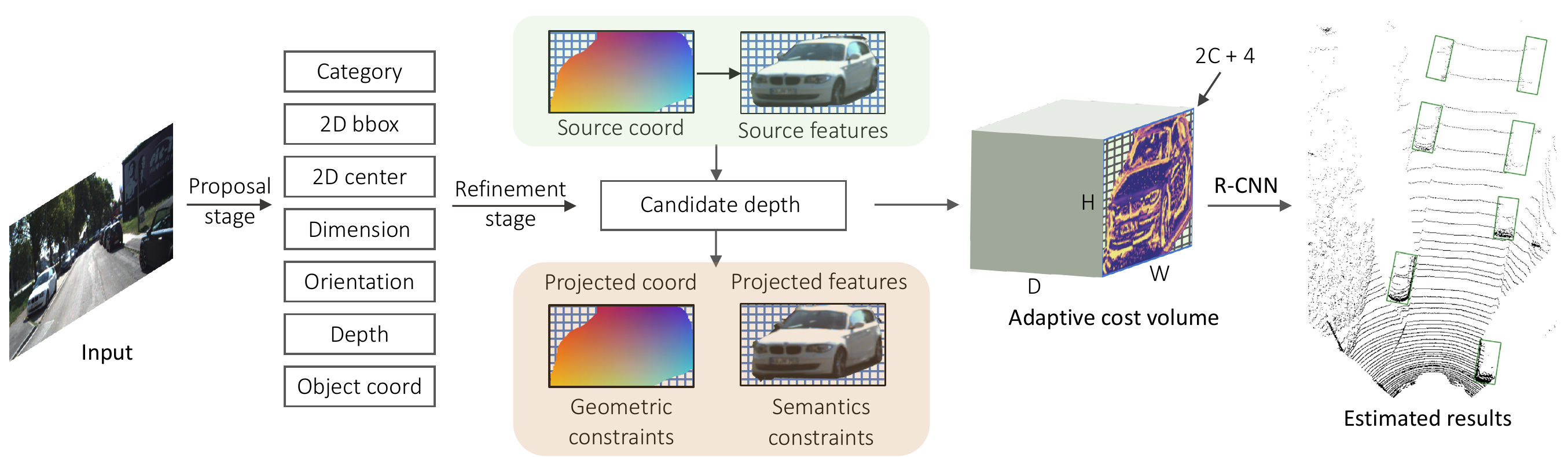}
    \caption{The visualization of our framework pipeline. In the proposal stage, we leverage a modified CenterNet to generate 3D bounding box proposals and estimate normalized object coordinate. Then, we construct the adaptive cost volume with the corresponding source and target coordinate with semantic features. Final bounding boxes are obtained by taking the cost volume to refine bounding box proposals. For visualization, we utilize the raw image to represent the learned semantic features. The attached image in the adaptive cost volume represents the error with the joint semantic and geometric energy function. The color from purple to yellow corresponds to the error value from 0 to 1. }
    \label{fig:overall}
\end{figure*}
\section{Background}

\subsection{Problem definition}
\label{sec:background}
In monocular 3D object detection, detectors are required to estimate the object dimension $(w, h, l)$, location of the object center $P_o = [x_o, y_o, z_o]^T$ and the corresponding yaw angle $\rho$. The estimated information also can be formulated as a rigid transformation matrix $[R_{o\rightarrow c}, P_o]$ that converts the point in the normalized object coordinate to the camera coordinate. The conversion matrix $R_{o\rightarrow c}$ is defined as follows:
\begin{align}
  R_{o\rightarrow c} =  \left(\begin{array}{ccc}
    \frac{w}{2} \cdot \cos(\rho) & 0 & \frac{w}{2} \cdot \sin(\rho) \\
    0  & \frac{h}{2} & 0 \\ 
    -\frac{l}{2} \cdot \sin(\rho)  & 0 & \frac{l}{2} \cdot \cos(\rho) \\ 
\end{array}\right).
\end{align}
And the rigid transformation for a point in the normalized object coordinate $^oP_i$ to the camera coordinate $P_i$ is formulated as:
\begin{align}
    \label{eq:coord}
    P_i = R_{o \rightarrow c}  {^oP_i} + P_o.
\end{align}

To recover the 3D information based on a 2D image, most of the monocular-based detectors~\cite{zhou2019objects, brazil2019m3d} first regress the projection location of $P_o$ in the image coordinate $p_0 = [u_o, v_o, 1]^T$ and then recover the 3D location by estimating the corresponding depth $z_o$.

As shown in Figure~\ref{fig:overall}, our detector first generates candidate proposals based on CenterNet~\cite{zhou2019objects}, which outputs the object classification, 2D bounding boxes, 2D projection location of the 3D center, object dimension, and yaw angle. For the depth estimation, we follow MonoDLE~\cite{monodle} and further estimate the standard deviation with Laplacian distribution during the training and inference stages. The loss function is formulated as follows:
\begin{align}
    \label{eq:depth}
    \mathcal{L}_{depth} = \frac{\sqrt{2}}{\sigma_z} \|\hat{z} - z^*\| + \log{\sigma_z},
\end{align}
where $\sigma_z$ denotes the  estimated standard deviation of depth, $\hat{z}$ and $z^*$ represent the estimated and ground truth depth, respectively.

\subsection{Geometry constraint with 2D-3D coordinate}
\label{sec:object_coord}
In this section, we introduce the way we build a pixel-level 2D-3D matching and the corresponding geometric constraint. 

\noindent{\textbf{Learning normalized object coordinate}}
Compared with the  camera coordinate, the normalized object coordinate is an easier learning objective due to its scale-invariant property under different locations and view angles~\cite{li2019multisensor, MonoRun}. Therefore, we convert the location of each pixel $i$ in the camera coordinate $P_i$ to the normalized object coordinate $^oP_i$. 
Then we add a branch in the CenterNet to estimate the location of each pixel in the normalized object coordinate.
For the pixel that has the correspondent lidar point, we convert the point cloud to the normalized object coordinate and directly minimize the $\ell_1$ distance between the estimated normalized object coordinate with the ground truth normalized object coordinate. 
For the pixel that do not has the correspondent lidar point, we adopt an unsupervised loss~\cite{MonoRun} that converts the estimated $^o\hat{P}_i$ to the image coordinate $\hat{p}_i$ and minimize the $\ell_1$ distance between it and $p_i$. 

\noindent{\textbf{Depth by solving 2D-3D constraint}}
With estimating the location of each point in the normalized object coordinate, a pixel-level energy function is built to represent the depth error of the bounding box. Specifically, we first project the estimated normalized object coordinate to the image coordinate based on the bounding box proposal:
\begin{align}
    \label{eq:coord_convert}
    \hat{p_i} =  \pi(K, \hat{R}_{o \rightarrow c} {^o\hat{P}_i} + \hat{P_o}),
\end{align}
where $\pi$ denotes the coordinate conversion from the camera to image coordinate and K is the camera intrinsic matrix. 
Then an energy function between the source location and the projected location of each pixel in the image coordinate is built:
\begin{align}
    E = \frac{1}{n}\sum_{i}^{n}\|p_i - \hat{p}_i\|,
\end{align}
where $n$ denotes the number of selected pixels in the bounding box. 
Traditional approaches~\cite{RTM3D, li2019multisensor, MonoRun} construct sparse or dense 2D-3D matching and recover depth by solving similar geometric-based energy functions.


\section{Approach}

As we introduced in Section~\ref{sec:intro}, depth recovery based on geometric constraints suffers from  indiscriminative 2D cues and the indirect optimization gap between 2D cue prediction and depth estimation. To alleviate these issues, we first propose a joint semantic and geometric energy function to enrich the geometric cue for depth reasoning. Based on the estimated 2D box and roi module~\cite{He_2017_ICCV}, we extract object level features with $F \in \mathcal{R}^{W \times H \times C}$ from the last feature extraction layer in CenterNet. For each pixel $i$, we utilize its origin location $p_i$ and the projected location $\hat{p_i}$ to sample the corresponding semantic features $F(p_i)$ and $F(\hat{p}_i)$. Bi-linear interpolation is adopted to ensure the sampling procedure differential. 
Through extracting the corresponding semantic features, we construct a joint semantic ($F(p_i)  \text{ \textit{vs.} }  F(\hat{p}_i)$) and geometric  ($p_i  \text{ \textit{vs.} } \hat{p}_i$) constraint.

\subsection{Refinement by adaptive 2D-3D cost volume}
\label{sec:method}
By incorporating the semantic features into the 2D-3D constraint, the energy function is more powerful than before. However, the  semantic features also lead to a non-convex energy function, making it unsolvable by fast linear solvers. Instead of adopting a complicated solving module,  we treat the built energy function as depth refinement features to indicate depth error. To provide effective refinement features, we construct an adaptive 4D cost volume that draws the error distribution of depth with the proposed energy function.

\noindent{\textbf{Adaptive Cost Volume}}
For each proposal bounding box, the 4D matching cost volume is built by concatenating the origin and projected semantic and geometric features in the image coordinate (with a size of $W\times H\times D\times 2(C+2)$, $H$: height of the roi feature, $W$: width of the ROI feature, $D$: number of the sampled depth, the first ``$2$'' is from the concatenation operation, $C$: the dimension of semantic features, the second ``$2$'': the dimension of pixels' location in the image coordinate). Specifically, the features contain the source and the projected location in the image coordinate $p$ and $\hat{p}$ and their corresponding  semantic features: $F(p)$and $F(\hat{p})$, where the $\hat{p}$ is determined by the candidate depth based on Equation~\ref{eq:coord_convert}.
Then we adopt a coordinate normalization to obtain location invariant geometric features for refinement:
\begin{align}
    ^op = \left( \begin{array}{ccc}
        \frac{1}{W} & 0 & -\frac{u_o}{W} \\
         0 & \frac{1}{H} & -\frac{v_o}{H} \\ 
    \end{array}\right) p^T,
\end{align}
where $[u_o, v_o]^T$ is the location of the projected center point in the image coordinate. 
As demonstrated in MonoDLE~\cite{monodle}, depth error changes with variant location, occlusions, \textit{etc}.
Hence, to adapt the cost volume with variant error distribution, the sampling size should be large enough to cover the potential ground truth. However, A too large sampling size would introduce high latency and memory occupation.
As a result, we adopt an adaptive sampling strategy that determines the size of the cost volume based on the uncertainty in depth estimation. Specifically, we leverage the estimated depth uncertainty $\sigma_z$ in the proposal stage to compute the sampling size.
During sampling, we first set the number of candidates with a fixed value $D$ and then determine the size of the depth grid $\Delta z$ based on $\sigma_z$: 
\begin{align}
    \Delta z =  \lambda \cdot \sigma_z,
\end{align}
where $\lambda$ is a pre-defined hyper-parameter.

\noindent{\textbf{Network architecture of the refinement module}}
Through modeling the error distribution, we design a refinement network that takes the distribution as refinement features for depth recovery. We first adopt several 2D convolution modules to extract features from the spatial space.
Then we insert 3D convolution networks to aggregate the features in the depth direction. 
After that, an average pooling layer and fully-connected networks with softmax activation function are employed to  integrate the features in the $H$ and $W$ direction and output final estimation. The output vector $\mathbf{\sigma_{d}} \in \mathcal{R^D}$ is processed with a soft arg-margin function~\cite{psmnet}  to compute the expectation for all candidate depth with probability:
\begin{align}
    \hat{z}_{refine} =  \hat{z} + \sum_{i = 1}^{D} + \Delta z \cdot \sigma^i_d \cdot (i - \frac{n}{2}),
\end{align}
where $\hat{z}$ is the estimated depth in the proposal stage and $\sigma^i_d$ denotes the estimated probability of the $i^{th}$ candidate depth. Compared with directly regressing depth value, the soft arg-margin operation with softmax function would encourage the model to learn discriminative features and select the optimal depth candidate. 
With refining the candidate depth $\hat{z}_{refine}$, the depth estimation loss based on Equation~\ref{eq:depth} is adopted to update the refinement network. 
Because the sampling operation in selecting semantic features is differentialable, the neural network can be trained to learn  suitable features for refinement. As the error map visualized in Figure~\ref{fig:overall},  the semantic representation ignores the texture-less regions and highlights the semantic regions.
During inference, the final 3D bounding boxes are obtained by the combination of the refined depth $\hat{z}_{refine}$ and the object dimension, yaw angle, and projection location estimated in the proposal stage.

\subsection{Overall pipeline}

The overall pipeline is visualized in Figure~\ref{fig:overall}. The proposal module is the modified CenterNet as described in Section~\ref{sec:background}. The refinement module is based on the cost volume described in Section~\ref{sec:method}. During training, the loss for optimizing the proposal module is the same as MonoDLE~\cite{monodle}. For the refinement loss, we filter the negative samples that the corresponding 2D IoU with ground truth is smaller than 0.5. During inference, we select candidate bounding boxes based on CenterNet and generate the final bounding boxes by combining the estimated category, dimension, yaw angle, projection location in the proposal stage and the estimated depth in the refinement stage.

\begin{table*}[htbp]
  \centering
  \caption{Comparison of the car category on the KITTI test set. We highlight the best results in \textbf{bold}. For the depth assisted-based approach (PatchNet~\cite{ma2020rethinking}, PCT~\cite{pct}), the inference time of depth estimator is from~\cite{ma2020rethinking}. Data split denotes the used data during training. ``Det'' denote the standard split for training 3D object detection and ``Eigen'' denotes the a set of unlabeled sequences in the KITTI dataset. PatchNet~\cite{ma2020rethinking} and PCT~\cite{pct} use eigen split to train a depth estimation model, MonoEF~\cite{MonoEF} uses it to train extrinsic estimation network and Kinemantic~\cite{kinematic} uses it to supervise the ego-motion network.}
    \begin{tabular}{l|c|cccccc|c} \hline
    \multirow{2} * {Setting}  & \multirow{2} * {Data split} & \multicolumn{3}{c}{3D (Test)} & \multicolumn{3}{c|}{BEV (Test)} & \multirow{2} * {Runtime (ms)}  \\ \cline{3-8}
       & &Easy & Mod & Hard & Easy & Mod & Hard & \\ \hline
    RTM3D (ECCV20)~\cite{RTM3D} & \multirow{9} * {Det} & 14.41 & 10.34 & 8.77 & 19.17 & 41.20 & 11.99 & 50 \\
    RAR-Net (ECCV20)~\cite{liu2020reinforced} & &  16.37 & 11.01 & 9.52 & 22.45 & 15.02 & 12.93 & - \\
    MonoDLE (CVPR21)~\cite{monodle} & & 17.23 & 12.26 & 10.29 & 27.94 & 17.34 & 15.24 & 40 \\
    M3DSSD (CVPR21)~\cite{M3DSSD} & &17.51 & 11.46 & 8.98  & 24.15 & 15.93 & 12.11 & - \\
    CaDDN (CVPR21)~\cite{CaDDN}&& 19.17 & 13.41 & 11.46 & 27.94 & 18.91 & 17.19  &  630 \\
    MonoRun (CVPR21)~\cite{MonoRun} & &19.65 & 12.3  & 10.58 & 27.94 & 17.34 & 15.24 & 70 \\ 
    MonoFlex (CVPR21)~\cite{MonoFlex} & & 19.94 & 13.89 & \color{blue}{12.07} & 28.23 & 19.75 & 16.89 & 30 \\
    Mono R-CNN (ICCV21)~\cite{decomp_shi} & & 18.36 & 12.65 & 10.03 & 25.48 & 18.11 & 14.10 & 70 \\
    AutoShape (ICCV21)~\cite{autoshape_liu} & &  \color{blue}{22.47} & \color{blue}{14.17} & 11.36 & \color{blue}{30.66} & \color{blue}{20.08} & 15.59 & 50 \\ \hline
    Kinemantic* (ECCV20)~\cite{kinematic} & \multirow{5} * {\makecell{Det +\\ Eigen}} &19.07 & 12.72 & 9.17  & 26.69 & 17.52 & 13.10 & - \\
    PatchNet (ECCV20)~\cite{ma2020rethinking}& & 15.68 & 11.12 & 10.17 & 22.97 & 16.86 & 14.97 &  488 \\
    MonoEF (CVPR21)~\cite{MonoEF} & &  21.29 & 13.87 & 11.71 & 29.03 & 19.70 & \textcolor{blue}{17.26} & 30 \\
    DFR-NET (ICCV21)~\cite{Zou_2021_ICCV} &  & 19.40 & 13.63 & 10.35 & 28.17 & 19.17 & 14.84 & 455 \\ 
    PCT (NeurIPS21)~\cite{pct} &  &21.00 & 13.37 & 11.31 & 29.65 & 19.03 & 15.92 & 487\\
    \hline
    MonoJSG&  Det& \textbf{24.69} &	\textbf{16.14} & \textbf{13.64} & \textbf{32.59} & \textbf{21.26} & \textbf{18.18} &42  \\ 
    \textit{Improvement} & - & \textit{+2.22} & \textit{+1.97} &  \textit{+1.54} & \textit{+1.93} & \textit{+1.18} &  \textit{+0.92} & - \\\hline
    \end{tabular}%
  \label{tab:kitti_test}%
\end{table*}




\section{Experiments}
\subsection{Experimental setup}
To demonstrate the efficacy of the proposed approach, we carry out experiments on both the KITTI~\cite{geiger2012we} and Waymo~\cite{waymo} monocular 3D object detection benchmarks.

\textbf{KITTI} dataset consists of 7,481 training and 7,396 testing images with annotating 80,256 3D bounding boxes. We follow 3DOP~\cite{chen20153d} and further split the training set into subsets with 3,712 images for training and 3,619 images for validation. Results with metrics of $AP|_{R40}$ in the space of 3D and BEV are reported. The bounding boxes are classified into three levels of difﬁculty: ``Easy'', ``Moderate'', and ``Hard'', determined by the height of the 2D bounding box, the object's occlusion, and truncation level. We train the model on three categories "Car", "Pedestrian" and "Cyclist" simultaneously and mainly report the results on the Car category.

\textbf{Waymo} open dataset is another large-scale autonomous driving dataset, which contains 1,150 video sequences collected from diverse driving environments. The official protocol splits the dataset into 798 training sequences, 202 validation sequences, and 150 test sequences. We follow  PCT~\cite{pct} and adopt the data from the front camera for monocular 3D object detection. For a fair comparison, we sample the images with every 3rd frame from the training sequences in the version of 1.2 (52,386 images) for training. We adopt the official evaluation tools~\cite{waymo} to calculate the mAP (mean average precision) and mAPH (mean average precision weighting by heading). Different from the KITTI dataset, Waymo separates the bounding boxes into two difficulty levels: "Level 1" and "Level 2" based on the number of lidar points contained in the bounding box. 

\noindent{\textbf{Implementation details}}
We follow the recent work~\cite{MonoFlex, autoshape_liu, MonoEF, M3DSSD} and adopt the commonly used CenterNet~\cite{zhou2019objects} with a modified DLA-34 backbone as the baseline detector. For the KITTI dataset, the input images are kept with the original resolution and pad to the size of 1280$\times$384 for training and inference. For the Waymo dataset, the input images are down-sampled to the size of 960$\times$640 to save computation time. We adopt the AdamW optimizer to train the model and set the initial learning rate as 4e-3. The network is initialized with ImageNet pre-trained weights and trained with 90 epochs on the KITTI dataset and 15 epochs on the Waymo dataset. During training, we only adopt random horizontal flip to augment the input image.

\begin{table}[htbp]
  \centering
  \caption{3D object detection results of Pedestrian and Cyclist on the KITTI test set.}
  \tabcolsep5pt
  \label{tab:ped_cyc}
    \begin{tabular}{l|ccc|ccc} \hline
  \multirow{2}[0]{*}{Method} & \multicolumn{3}{c|}{Pedestrian} & \multicolumn{3}{c}{Cyclist} \\ 
  & Easy & Mod & Hard &  Easy & Mod & Hard  \\  \hline
    MonoPair~\cite{monopair} & 10.02 & 6.68  & 5.53  & 3.79  & 2.12  & 1.83 \\
    MonoFlex~\cite{MonoFlex} & 9.43  & 6.31  & 5.26  & 4.17  & 2.35  & 2.04 \\
    Autoshape~\cite{autoshape_liu} & 5.76  & 3.74  & 3.03  & \textbf{5.99}  & 3.06 & \textbf{2.70} \\
    MonoRun~\cite{MonoRun} & 10.88 & 6.78  & 5.83  & 1.01  & 0.61  & 0.48 \\
    MonoJSG  & \textbf{11.02} & \textbf{7.49}  & \textbf{6.41}  & 5.45  & \textbf{3.21}  & 2.57 \\ \hline
    \end{tabular}%
\end{table}%

\subsection{Benchmark evaluation}

\subsubsection{Results on the KITTI test set}
In Table~\ref{tab:kitti_test}, we present the experimental results of our detectors and the other state-of-the-art methods on the KITTI test set.
We draw the following observations: (1) Our approach achieves the best performance of the car class on six different metrics. Compared with the second-best approach, our approach outperforms them with the ratio of 9.89\%, 13.90\%, and 12.75\% on the 3D detection task and 6.29\%, 5.88\%, and 5.33\% on the BEV detection task with the ``Easy'', ``Moderate'', and ''Hard'' setting, respectively.  Furthermore, the designed modules in our detector are lightweight, making it applicable for autonomous driving systems. (2) Compared with the geometric constraint based methods (\textit{i.e.,} RTM3D~\cite{RTM3D}, AutoShape~\cite{autoshape_liu}, and MonoRun~\cite{MonoRun}), our approach keeps a similar inference time but achieves much better 3D detection performance, showing the effectiveness of incorporating semantic features for depth recovery. It is important to note that the second-best method AutoShape~\cite{autoshape_liu} adopts extra CAD models to learn the normalized object coordinate, where our approach is orthogonal to the proposed approach.  
(3) Furthermore, the depth-assisted based approaches require an extra depth estimation model, leading to a heavy computation burden~\cite{ma2020rethinking} during inference. In contrast, our detector not only keeps a lightweight framework but also achieves better performance.

In Table~\ref{tab:ped_cyc}, we further display the experimental results with Pedestrian and Cyclist class on the KITTI test set. Our method obtains the best performance in the Pedestrian class and achieves comparable performance with the best approach Autoshape~\cite{autoshape_liu} in the Cyclist class. 
It is worth noting that the number of the annotated instances in the two categories is small (Pedestrian with 4,487 and Cyclist with 1,627 \textit{vs.} Car with 28,742 in the training set), which may introduce performance fluctuation. 

\begin{figure}[htbp]
    \centering
    \includegraphics[width = 0.49\textwidth]{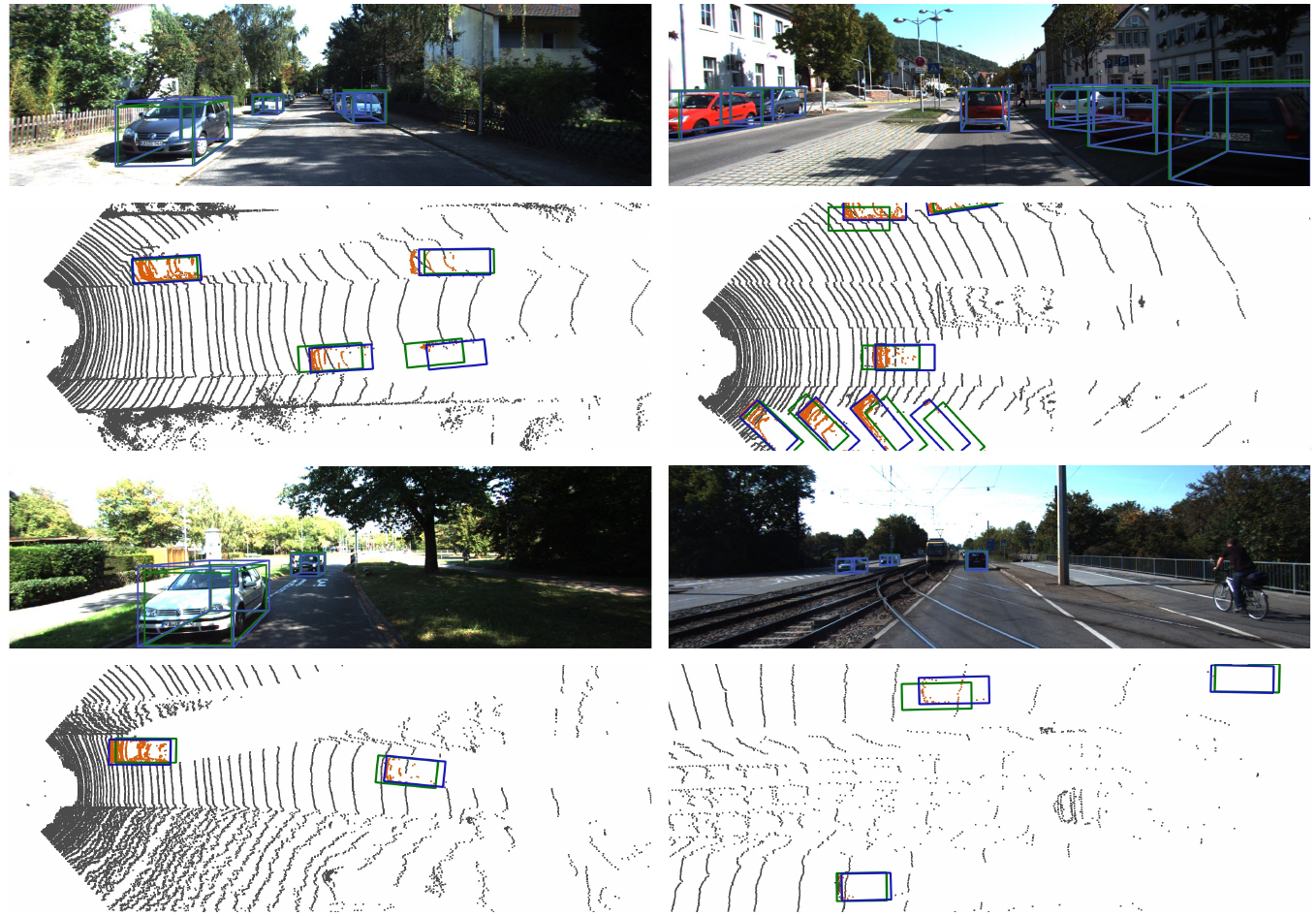}
    \caption{Qualitative results of before and after refinement on the KITTI dataset. ``\textcolor{green}{Green box}'' and ``\textcolor{blue}{Blue box}'' represent the results before and after refinement. }
    \label{fig:qualitative}
\end{figure}

\begin{table*}[htbp]
\small
 \caption{Experimental results of the Car category on the Waymo validation set. We adopt the Metrics with mAP and mAPH under the IoU threshold of 0.7 and 0.5, respectively. ``Level 1'' denotes the evaluation of the bounding boxes that contain more the 5 lidar points. ``Level 2'' denotes the evaluation of all of the bounding boxes. } 
 
 \centering
 \tabcolsep4pt
 \begin{tabular}{ll|cccc|cccc}
  \hline

 \multirow{3} * {Setting} & \multirow{3} * {Method} & \multicolumn{8}{c}{3D mAP / 3D mAPH}   \\ \cline{3-10}
 & & \multicolumn{4}{c|}{IoU = 0.7} & \multicolumn{4}{c}{IoU = 0.5}\\ 
 &  & Overall & 0 - 30m          & 30 - 50m          & 50 - $ \infty$        & Overall & 0 - 30m          & 30 - 50m          & 50 - $ \infty$     \\
  \hline

 \multirow{4}*{Level 1} &  PatchNet (ECCV20)~\cite{ma2020rethinking}  &  
      0.39/0.37 &  1.67/1.63  & 0.13/0.12 & 0.03/0.03 & 2.92/2.74 &  10.03/9.75  & 1.09/0.96 & 0.23/0.18 \\
   &  PCT (NeurIPS21)~\cite{pct}   
     & 0.89/0.88 &  3.18/3.15  & 0.27/0.27 & 0.07/0.07  & 4.20/4.15 &  14.70/14.54  & 1.78/1.75 & 0.39/0.39 \\
    & Baseline & 0.78/0.76 &3.80/3.73 & 0.49/0.48 & 0.08/0.07 & 4.59/4.47 & 18.35/17.93 & 3.16/3.09 & 0.74/0.10  \\
    & MonoJSG & \textbf{0.97/0.95} & \textbf{4.65/4.59} & \textbf{0.55/0.53} & \textbf{0.10/0.09} &  \textbf{ 5.65/5.47} & \textbf{20.86/20.26} & \textbf{3.91/3.79} & \textbf{0.97/0.92}  \\
    \hline
  \multirow{4}*{Level 2} &  PatchNet (ECCV20)~\cite{ma2020rethinking}  &  
  0.38/0.36 & 1.67/1.63 &  0.13/0.11 & 0.03/0.03 & 2.42/2.28 & 10.01/9.73 & 1.07/0.94 & 0.22/0.16 \\
    & PCT (NeurIPS21)~\cite{pct}  & 0.66/0.66 & 3.18/3.15 & 0.27/0.26 & 0.07/0.07 & 4.03/3.99 & 14.67/14.51 & 1.74/1.71 & 0.36/0.35 \\
     & Baseline & 0.74/0.72 & 3.79/3.72 & 0.48/0.47 & 0.07/0.07  & 4.34/4.22 & 18.33/17.87 & 3.07/3.00 & 0.65/0.63 \\ 
    & MonoJSG & \textbf{0.91/0.89} & \textbf{4.64/4.65} & \textbf{0.55/0.53} & \textbf{0.09/0.09} & \textbf{5.34/5.17} & \textbf{20.79/20.19 } & \textbf{3.79/3.67} & \textbf{ 0.85/0.82} \\
    \hline
 \end{tabular}
 \label{tab:waymo}
  \vspace{-1.1em}
\end{table*}

\subsubsection{Results on the Waymo validation set}
In Table~\ref{tab:waymo}, we compare the proposed approach with recent top-performed approaches~\cite{ma2020rethinking, pct} on the Waymo validation set. We report the evaluation results of the Car category with two different IoU thresholds (0.7 and 0.5). Similar to the observation on the KITTI dataset, our approach yields consistent improvement over state-of-the-art approaches. Compared to the second-best approach PCT~\cite{pct}, we improve them over 5.61\% and 26.42\% on 3D mAP with IoU threshold of 0.7 and 0.5 respectively. We display the detailed qualitative results in the supplementary.


\begin{figure*}
    \centering
    \includegraphics[width=\textwidth]{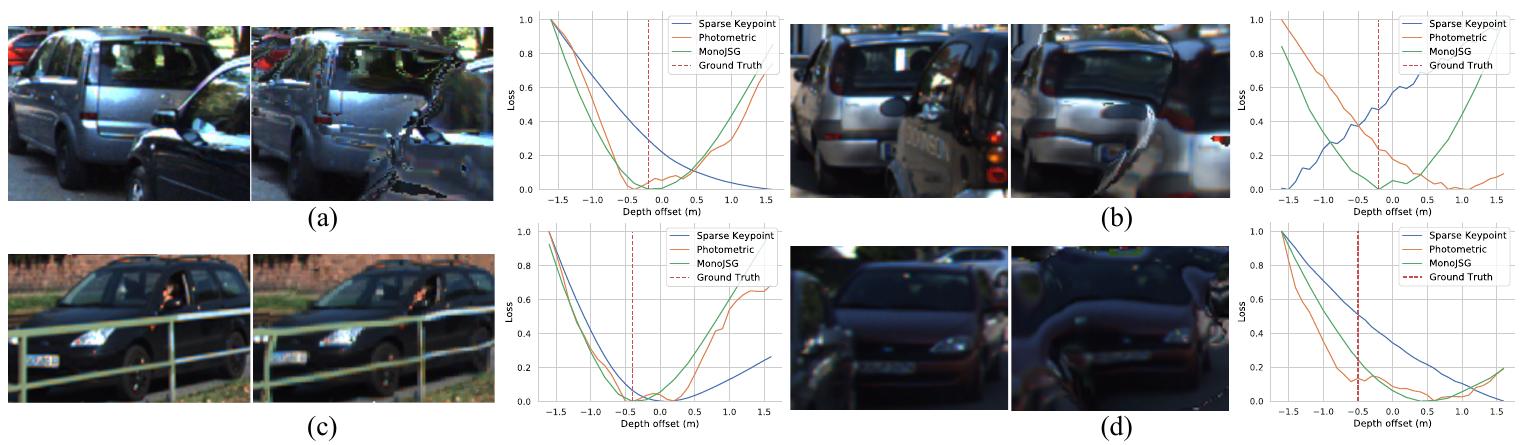}
    \caption{From left to right: Visualization of the origin images, the projected images and the loss distribution of the energy function with different 2D cues. The loss distribution is drawn from the range of [-1.6m to 1.6m] around the estimated depth. ``Sparse Keypoint'' denotes the eight bounding box corners. ``Photometric'' denotes the raw image.}
    \label{fig:proj_distribution}
\end{figure*}

\begin{table}[htbp]
  \centering
  \caption{Experimental results of constraint-based and refinement-based approaches with different input features on the KITTI validation set.  The constraint based approaches adopt the depth that achieves minimal error in the constraint as final results. ``Sparse Geo'' and ``Dense Geo'' denote using the eight bounding box corners~\cite{RTM3D, km3dnet} and all the pixels in the objects~\cite{MonoRun} to build the constraint. MonoJSG denotes using the pixel-level joint semantic and geometric features for building the constraint.}
    \label{tab:constraint_refine}
    \begin{tabular}{l|ccc}\hline
    Setting  & Easy  & Mod   & Hard \\ \hline
    Baseline    & 20.5  & 14.2  & 12.0 \\ 
    Sparse Geo Constraint & 18.9  & 13.1 & 10.8 \\ 
    Dense Geo Constraint &  22.3  & 15.7  & 13.3 \\ \hline
    MonoJSG Constraint & 24.3 &17.0  & 14.5     \\
    MonoJSG Refine & \textbf{26.4}  & \textbf{18.3}  &\textbf{15.4}\\ \hline
    \end{tabular}%
    \vspace{-3mm}
\end{table}%

\subsection{Ablation study}
In Table~\ref{tab:constraint_refine} , we compare the different ways of recovering depth, and validate the improvement of using semantic representations for depth recovering on the KITTI validation set.  Surprisingly, the 3D bounding boxes estimated by sparse geometry constraint are even worse than the baseline model. On the opposite, the dense constraint improves the baseline model with 1.8\% mAP in the Easy setting. The performance degradation may come from the inaccurate keypoint localization. Compared with the dense geometric constraint, MonoJSG provides a data-driven fashion to learn the visual cue, leading to better detection performance. With the designed refinement module, MonoJSG achieves the best performance among variant settings. 

\begin{table}[htbp]
  \centering
  \caption{Ablation study of different input features in the refinement module. ``Geo'', ``Photo'' and ``Semantic'' denote the location of each pixel in the image coordinate, raw photo feature and the learned semantic representation respectively.}
    \begin{tabular}{cccc|ccc} \hline
   \# & Geo & Photo & Semantic& Easy & Mod & Hard \\ \hline
    0&     &   &    &    20.5   &  14.2     &12.0  \\ 
    1&\cmark      &       &       &   22.4    &  15.6 & 13.3  \\
     2&\cmark     &  \cmark& & 24.3      &  16.2     & 13.4 \\
    3&\cmark  &  & \cmark & 26.3 & \textbf{18.4} & \textbf{15.4} \\
    4&\cmark & \cmark & \cmark & \textbf{26.4} & 18.3 & \textbf{15.4} \\  \hline
  \label{tab:refine}
    \end{tabular}%
\end{table}%

\noindent{\textbf{Different features for refinement}}
In Table~\ref{tab:refine}, we further present the comparison of using different features for refinement. 
Compared with only using the geometric features, the photometric and semantic features based approaches can utilize the provided visual cue to identify the discriminative region for refinement, which yields better detection results. 
Furthermore, the comparison between experiments ``2'' and ``3'' demonstrates the effectiveness of leveraging the learned representation for refinement. 
By comparing the experiment ``4'' and ``5'', we observe a limited improvement of incorporating photometric features to MonoJSG. This is because that the semantic features are extracted from the raw image and act as a similar role in refinement.
\begin{table}[htbp]
  \centering
  \caption{Ablation of the sampling operation in the proposed cost volume.} 
   \label{tab:volume}
    \begin{tabular}{cc|ccc} \hline
  Sample size & Strategy & Easy & Mod & Hard \\ \hline
    \multirow{2}*{8} & Uniform & 23.6 & 16.9 & 13.8 \\
     &  Adaptive & 24.7 & 17.2 & 14.4\\ \hline
    \multirow{2}*{32} & Uniform & 25.2 & 17.3 & 14.7 \\
     &  Adaptive & \textbf{26.4} & \textbf{18.3} & \textbf{15.4} \\ \hline
     \vspace{-4mm}
    \end{tabular}%
\end{table}%

\noindent{\textbf{Different sampling strategies in the cost volume}}
To validate the effectiveness of the proposed adaptive sampling strategy in the cost volume, we compare it with the commonly used uniform sampling strategy on the KITTI validation set. We adopt the size  of 8 and 32 to sample candidate depth. As illustrated in Table~\ref{tab:volume}, the adaptive sampling consistently outperforms the uniform sampling in the size of 8 and 32. With more candidate depth, the cost volumes with the size of 32 achieve better performance than the size of 8, while it also increases 8x memory occupation in the refinement module. 

\subsection{Qualitative results}
In Figure~\ref{fig:qualitative} and~\ref{fig:proj_distribution}, we provide the qualitative results of our detector on the KITTI datasets. Compared with the sparse geometric error and photometric error, the loss landscape of our MonoJSG is more robust. As illustrated, the loss curve of MonoJSG is more convex and the corresponding depth of minimum value is near to the ground truth. On the opposite, as illustrated in instances (a) and (b), the sparse keypoint based constraint fails in the occlusion situation. The instance in Figure~\ref{fig:proj_distribution} (d) shows a failure case that the neural network fails to estimate accurate object coordinate in the low-visibility situation, which also contributes to inaccurate geometric and semantic cues for depth recovery. However, the low-visibility is a typical problem in computer vision, which could be addressed in the image pre-processing stage.

\section{Conclusion and Limitations}
We have presented MonoJSG, a refinement-based monocular 3D object detection framework for autonomous driving scenarios.
Benefits from both the powerful feature representation from DNN and the pixel-level visual cues from 2D-3D constraint, MonoJSG effectively mitigates the object depth error, leading to state-of-the-art results on the KITTI and Waymo datasets.

Similar to other geometric constraint based approaches, the accuracy of our 2D-3D constraint is based on the estimated object dimension and yaw angle. Although they are easier to estimate than object depth, they may fluctuate in the low-visibility or far away regions. Modeling their distribution and incorporating their uncertainty to the constraint may alleviate the limitation.

\noindent{\textbf{Potential impacts}}
This work studies monocular 3D object detection in autonomous driving. The potential security risk of this work is that the localization error in the model may mislead the following motion planning, which may lead to traffic accidents.  
{\small
\bibliographystyle{ieee_fullname}
\bibliography{egbib}
}







\end{document}